\documentclass[10pt,twocolumn,letterpaper]{article}

\usepackage{cvpr}              

\usepackage[dvipsnames]{xcolor}

\definecolor{cvprblue}{rgb}{0.21,0.49,0.74}
\usepackage[pagebackref,breaklinks,colorlinks,citecolor=cvprblue]{hyperref}

\usepackage[group-separator={,}]{siunitx}
\usepackage{subcaption}
\usepackage{amssymb,amsthm,amsmath,array}
\usepackage[linesnumbered, ruled, vlined]{algorithm2e}
\usepackage{svg}
\usepackage{hyperref}
\usepackage{algpseudocode}
\usepackage{multirow}
\usepackage[accsupp]{axessibility}
\algnewcommand\algorithmicLineComment[1]{\color{blue} /* #1 */ \color{black}}
\newcommand{\multicomment}[1]{}

\def\paperID{17} 
\def\confName{CVPR}
\def\confYear{2024}

\def\titlename{Multi-Objective Hardware Aware Neural Architecture Search using Hardware Cost Diversity}
\def\methodname{MO-HDNAS}
\title{\titlename}

\author{Nilotpal Sinha, Peyman Rostami, Abd El Rahman Shabayek, Anis Kacem, Djamila Aouada\\
{\tt\small \{nilotpal.sinha, peyman.rostami, abdelrahman.shabayek, anis.kacem, djamila.aouada\}@uni.lu} \\
SnT, University of Luxembourg\\
}

\begin{document}

\maketitle
\begin{abstract}
    Hardware-aware Neural Architecture Search approaches (HW-NAS) automate the design of deep learning architectures, tailored specifically to a given target hardware platform.
    Yet, these techniques demand substantial computational resources, primarily due to the expensive process of assessing the performance of identified architectures.
    To alleviate this problem, a recent direction in the literature has employed representation similarity metric for efficiently evaluating architecture performance.
    Nonetheless, since it is inherently a single objective method, it requires multiple runs to identify the optimal architecture set satisfying the diverse hardware cost constraints,
    thereby increasing the search cost. 
    Furthermore, simply converting the single objective into a multi-objective approach results in an under-explored architectural search space.
    In this study, we propose a Multi-Objective method to address the HW-NAS problem, called \methodname, to identify the trade-off set of architectures in a single run with low computational cost.
    This is achieved by optimizing three objectives: maximizing the representation similarity metric, minimizing hardware cost, and maximizing the hardware cost diversity.
    The third objective, \ie hardware cost diversity, is used to facilitate a better exploration of the architecture search space.
    Experimental results demonstrate the effectiveness of our proposed method in efficiently addressing the HW-NAS problem across six edge devices for the image classification task.
\end{abstract}

\section{Introduction}
\label{section:intro}
    Advancements in deep learning systems have brought about a revolutionary impact on various domains, particularly in computer vision~\cite{simonyan2014very, sermanet2013overfeat, zamani2021elliptical,rostami2022deep, perez2021detection, garcia2021lspnet, 9620184}, natural language processing \cite{collobert2011natural, wu2016google, devlin2018bert}, and more. These remarkable achievements were made possible by the creation of meticulously designed architectures that are specifically tailored for individual tasks.

    \begin{figure}[t]
    \centering
    \includegraphics[width=0.48\textwidth]{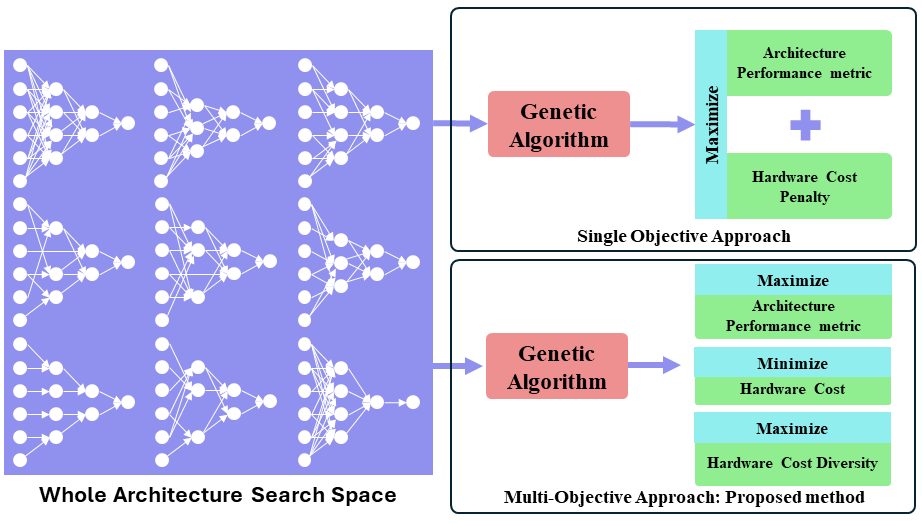}
    \caption{An illustration of the difference between 
    a single objective approach to HW-NAS problem and our proposed method \methodname.}
    \label{fig:proposed_method}
\end{figure}
    
    In response to the growing need for more advanced architectures, researchers have turned their focus towards developing algorithms that can effectively explore the extensive space of neural architectures. These algorithms, collectively known as Neural Architecture Search (NAS)~\cite{elsken2018neural, zoph2017neural, zoph2018learning}, are specifically designed to discover the most optimal architecture for a given task.
    \begin{figure*}[t]
        \centering
        \includegraphics[width=\linewidth]{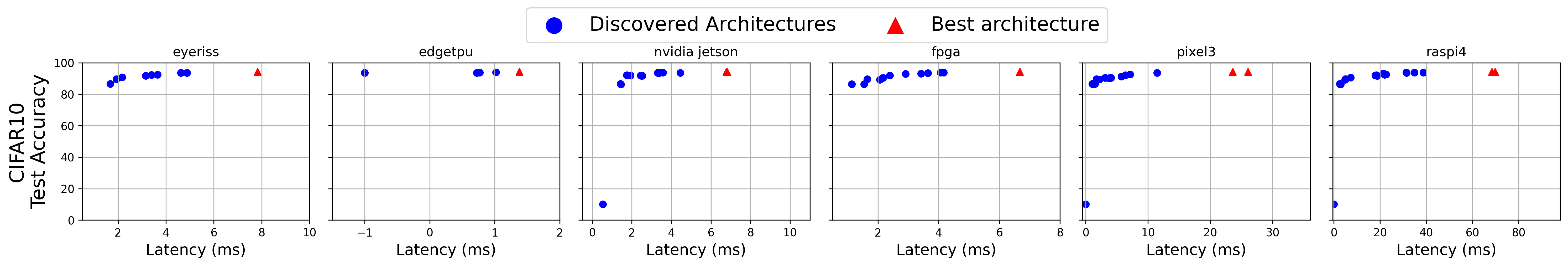}
        \caption{Results of a naive conversion from a single objective to a multi-objective NAS with two objectives: maximize representation similarity and minimizing device latency. It fails to identify the best architecture within the search space.
        The architecture search is performed in the search space, NAS-Bench-201~\cite{Dong2020NAS-Bench-201} on CIFAR10 dataset. More details about the search space are given in Section~\ref{subsection:search_space}.}
        \label{fig:2objs_problem}
    \end{figure*}
    
    The rise in the utilization of edge devices, characterized by low energy consumption, necessitated adaptations to NAS algorithms to incorporate performance considerations from the particular hardware being employed. These customized NAS algorithms are referred to as Hardware-aware Neural Architecture Search (HW-NAS)~ \cite{ijcai2021p592, benmeziane2021comprehensive}.
    While NAS focuses on finding the optimal architecture for a specific task, HW-NAS aims to find architectures with minimal trade-offs between task performance and targeted hardware cost.
    However, HW-NAS algorithms face a bottleneck due to the extensive time required for evaluating the architecture performance metrics within the search space~\cite{real2019regularized, zoph2018learning}.
    This challenge has led to the development of methods that utilize a \textit{supernet-based} solution~\cite{liu2018darts2, cai2019once, cai2018proxylessnas}, treating all architectures in the search space as sub-networks of the supernet.
    While employing this strategy reduces computational cost, it compromises architecture search performance due to inaccurate performance estimations by the supernet.

    To address the mentioned challenge,~\cite{sinha2024hardware} has recently proposed the use of a \textit{representation similarity metric}~\cite{kornblith2019similarity, zheng2022neural}, which significantly reduced the search cost while finding the best matching architecture under a single hardware cost constraint.
    This was achieved by using the \textit{single objective} of maximizing the representation similarity metric with respect to a reference model, while penalizing the search whenever the given architecture constraint is not satisfied (illustrated in Figure~\ref{fig:proposed_method}). 
    However, if multiple different constraints need to be satisfied, the search cost adds up as the algorithm must run multiple times to fulfill each one.
    Additionally, the naive conversion of the single objective method of~\cite{sinha2024hardware} to a \textit{multi-objective} one with two objectives (\ie  maximizing representation similarity metric and minimizing hardware cost) fails to identify the best architecture.
    In this regard, Figure~\ref{fig:2objs_problem} illustrates the hardware costs, measured in terms of device latency, of the set of architectures discovered after performing the multi-objective architecture search.
    It is evident from the figure that the architectures discovered through the architecture search do not exhibit similar performance (test accuracy) to the best architecture found within the search space.
    This failure is attributed to the high hardware cost of the best architecture, contradicting the second objective aimed at minimizing hardware costs.
    Note that best architecture in the figure refers to the architecture with the highest accuracy in the search space.

    \begin{figure*}[t]
    \centering
    \includegraphics[width=\linewidth]{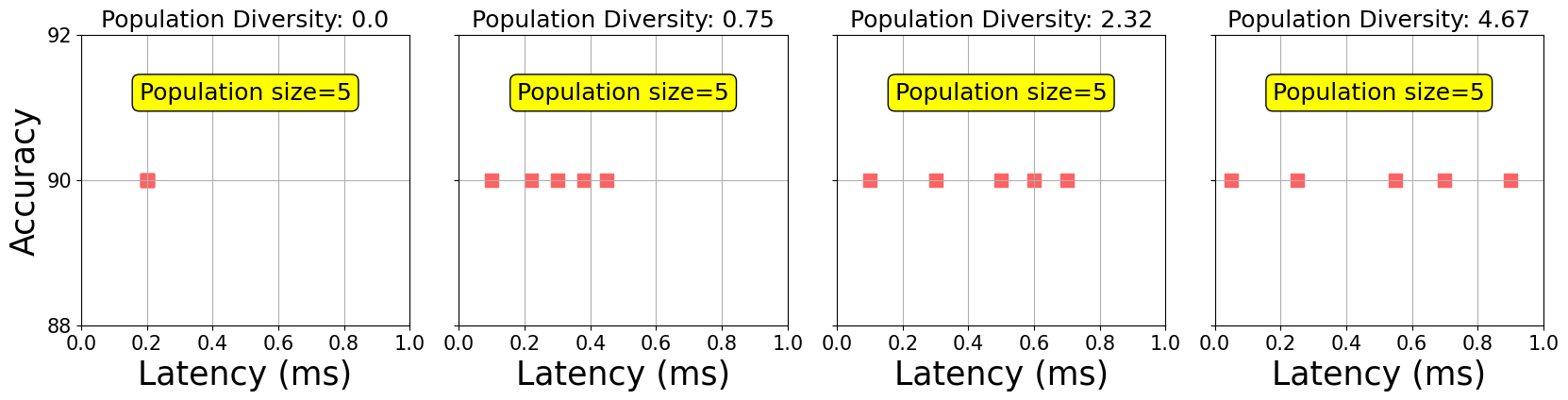}
    \caption{A depiction of the accuracy (y-axis) of all architectures against their respective hardware costs measured in terms of latency (x-axis), in a population size of 5.
    As the value of diversity increases, the architectures in the population exhibit a spread in hardware costs along the latency axis.
    }
    \label{fig:max_diversity}
\end{figure*}

    To address these challenges, we propose a Multi-Objective method to address HW-NAS called \titlename$\ $(\methodname). Our approach aims to  identify a set of high-performing architectures with diverse hardware costs in a single run.
    It achieves this goal by optimizing three objectives (illustrasted in Figure~\ref{fig:proposed_method}): (\textit{1}) Maximizing the representation similarity metric. (\textit{2}) Minimizing hardware cost. (\textit{3}) Maximizing hardware cost diversity.
    Our contributions can be summarized as follows:
        \begin{itemize}
        \item We generalize the single objective HW-NAS framework proposed in~\cite{sinha2024hardware} to a multi-objective one in order to address the issue of increased search cost when multiple hardware cost constraints are present.
        \item We propose a \textit{hardware cost diversity} term aimed at encouraging the consideration of architectures with diverse hardware costs.
        This allows the search algorithm to explore architectures with higher hardware costs, as high-performing architectures typically tend to have higher hardware cost requirements.
        \item The robustness of the proposed method is demonstrated on six different edge devices for classification tasks.
    \end{itemize}
    

\section{Related Works}
\label{section:related_work}
    Any NAS method, as described in  \cite{elsken2019neural}, consists of three key components: \textit{search space}, \textit{search strategy}, and \textit{performance estimation}.
    The search space generally outlines the potential architectures that can be theoretically represented. 
    Performance estimation involves assessing the expected performance of a neural architecture for a specified task.
    The search strategy dictates the approach used to explore the defined search space, utilizing architecture performance estimation to find the optimal architecture.
    It involves techniques such as reinforcement learning (RL)-based methods \cite{zoph2018learning, pmlr-v80-pham18a}, evolutionary algorithm (EA)-based methods 
    \cite{real2019regularized, sinha2021evolving, sinha2022neural, sinha2022novelty, 10.1162/evco_a_00331}, and gradient-based methods \cite{liu2018darts2, Zela2020Understanding}.

    Hardware-aware NAS (HW-NAS) is a specialized version of NAS aimed at identifying the optimal architecture tailored for a specific task and target device.
    HW-NAS typically involves addressing multiple objectives, such as maximizing the architecture performance metric while minimizing the associated hardware cost for the target hardware~\cite{ijcai2021p592}. 

    Addressing the challenge of multiple objectives can be pursued through two distinct approaches~\cite{benmeziane2021comprehensive}. The first method entails converting the multiple objective problem into a single objective one and solving the latter instead. This can be achieved via rejection sampling~\cite{cai2019once}, which eliminates any architecture that fails to meet the hardware cost constraint during the search process. However, rejection sampling is susceptible to the halting problem, as indicated by~\cite{sinha2024hardware}, particularly when it rejects all candidate architectures for failing to meet a low hardware cost constraint. An alternative solution to rejection sampling involves employing a penalty term that reduces the performance metric of an architecture whenever it does not satisfy the hardware cost constraint. Yet, this latter solution suffers from high computational cost. Specifically, when multiple hardware constraints are present, the same single objective problem  has to be solved multiple times to accommodate all hardware constraints. 
    The second method to address multiple objectives in HW-NAS employs techniques to identify the pareto optimal solutions~\cite{ying2020neural, chu2020multi}. 
    Pareto optimal solutions are those that cannot be improved in one objective without compromising at least one other objective.
    For instance, improving the accuracy of an architecture may require increasing network parameters, thereby elevating the hardware cost.  
    The pareto approach effectively tackles the elevated search cost issue linked to the single-objective relaxation.
    Hence, it will be exploited in this work as it provides a set of architectures (pareto optimal set) in a single run, in contrast to multiple runs required by the single objective approach. 
    Our method also ensures that these architectures have diverse hardware costs.
    This stands in contrast to previous multi-objective methods~\cite{ying2020neural, chu2020multi}, where the diversity of architecture hardware cost was not considered as one of the objectives.
\section{Proposed Method}
\label{section:proposed_method}

\subsection{Search Method}
    Our proposed architecture search method employs a meta-heuristic optimization technique falling under the category of \textit{genetic algorithms}~\cite{goldberg2013genetic}. These algorithms have demonstrated their effectiveness in addressing the NAS problem~\cite{sinha2021evolving, sinha2022novelty, sinha2022neural, sinha2024hardware}.
    They mimic biological adaptation to find optimal solutions in non-differentiable spaces.
    Starting with an initial population of random neural network architectures, the algorithm iteratively updates/evolves the population, ensuring that the new population $\mathcal{P}$ consists of better performing architectures as compared to previous one.  After running the algorithm for a certain number of iterations/generations, the best architecture in the current population is returned as the final solution.
    
    To solve the multi-objective problem of HW-NAS, 
    we employed a popular variant of the genetic algorithm called NSGA-II~\cite{deb2002fast}.
    It is a well-known \textit{Pareto-based Multi-objective Evolutionary Algorithm (MOEA)}, where selection of individuals is based on \textit{Pareto Efficiency}.
    In this context, a solution that outperforms others in all objectives is termed ``\textit{non-dominated}".
    Conversely, one that is inferior to others in at least one objective is consistently labeled as ``\textit{dominated}".
    During the selection phase, solutions undergo a sorting process using non-dominated sorting and crowding distance.
    This technique has been previously employed in the NAS~\cite{lu2019nsga, sinha2022novelty} literature, offering a suitable solution for optimizing a neural network architecture based on various objectives.
    
\subsection{Problem Formulation}
\label{subsection:problem_formulation}
    Let $\alpha^*$ denote a pre-trained \textit{reference model} with a desired performance metric (\eg  accuracy for classification task).
    Also, let $\mathcal{A}$ be the architecture search space in which  NAS is performed with $\alpha$ denoting an architecture in the search space.
    Further, let $\Psi(.)$ denote the function that measures the hardware cost (\eg latency).
    Formally, the multi-objective hardware-aware architecture search problem can be written as:
    \begin{align}
	\label{eq:problem}
        \max_{\alpha \in \mathcal{A}} \quad & \phi(\alpha^*, \alpha), \notag\\
        \min \quad &  \Psi(\alpha), \\
        \max \quad &  \chi(\alpha, \mathcal{P}). \notag
    \end{align}
    
    This formulation involves solving for three objectives, including:
    \begin{enumerate}
        \item \textit{Maximizing performance similarity metric}, $\phi(\alpha^*, \alpha)$: Finding an architecture $\alpha$ in the search space with similar performance to the reference model $\alpha^*$.
        More specifically, the performance similarity metric calculates the mutual information between hidden layer representation of an architecture and that of the reference model. In other words,
        \begin{eqnarray}\label{eq:rmi}
            \phi(\alpha^*, \alpha) =
            \sum_{i=1}^{L} I(X^{i*}, X^{i}), 
        \end{eqnarray}
        \noindent where $X^{1*}, X^{2*}, .., X^{L*}$ and $X^{1}, X^{2}, .., X^{L}$ represent the random variables of feature maps in each layer of $\alpha^*$ and $\alpha$, respectively.
        More details are available  in~\cite{sinha2024hardware, zheng2022neural}.
        \item \textit{Minimizing hardware cost}, $\Psi(\alpha)$: Finding an architecture $\alpha$ with minimum hardware cost.

        \item \textit{Maximizing hardware cost diversity}, $\chi(\alpha, \mathcal{P})$: Maximizing the diversity of the architecture $\alpha$ in terms of the hardware cost, \ie $\chi(\Psi(\alpha, \mathcal{P}))$, as will be discussed in Section~\ref{subsection:hardware_diversity}.
        Note that $\mathcal{P}$ refers to the current generation population of architectures.
    \end{enumerate}    
    

\subsection{Hardware Cost Diversity}
\label{subsection:hardware_diversity}
    For the current generation population $\mathcal{P}$, the hardware cost diversity term for each architecture $\alpha$ is calculated as 
    \begin{align}
	\label{eq:hardware_diversity}
        \chi(\alpha, \mathcal{P}) = \sum_{\alpha^{\dagger} \in \mathcal{P}} (\Psi(\alpha) - \Psi(\alpha^{\dagger}))^2.
    \end{align}

    \noindent This formulation measures the difference between the hardware cost of a given architecture $\alpha$, and those of the remaining architectures $\alpha^{\dagger}$ in the given population $\mathcal{P}$.
    Maximizing this term leads to a population characterized by architectures with  diverse hardware costs. This is illustrated in Figure~\ref{fig:max_diversity} which plots the hardware costs of architectures in a population of size five and the impact of the diversity term. This allowed the discovery of architectures with lower latency that preserve the same level of accuracy.
    
    To measure the diversity of the population, we introduce a term called \textit{population diversity}, $\bar{\chi}(\mathcal{P})$, formalized as
    \begin{align}
	\label{eq:avg_hardware_diversity}
        \bar{\chi}(\mathcal{P}) = \frac{1}{N}\sum_{\alpha \in \mathcal{P}} \chi(\alpha, \mathcal{P}),
    \end{align}
    where $N$ is the population size. It is worth mentioning that
    $\bar{\chi}(\mathcal{P})$ measures the average hardware cost diversity of architectures within the population.
    
    The leftmost plot in Figure~\ref{fig:max_diversity} shows the population of architectures with the same hardware cost, consequently resulting in the population diversity term being zero.
    As we progress to the right on the plots in the figure, we observe an increase in the population diversity term.
    This ensures that the search algorithm explores the architecture search space, encompassing architectures with varying hardware costs.
    Please note that the hardware cost used in Figure~\ref{fig:max_diversity} represents latency.
    However, the proposed method is agnostic to the specific type of hardware cost utilized for the architecture search.
   
    \begin{algorithm}[t]
		\caption{\methodname}	
		\label{algo}
		\SetAlgoLined
		\KwIn{Reference model $\alpha*$, Search space $\mathcal{A}$, Total generations $N_{gen}$, Population size $N_{pop}$, training epochs $N_{train}$}
		\KwOut{Pareto optimal front of architectures, $P_{optimal}$}
		$\mathcal{P} \gets $ Initialize population for NSGA-II algorithm\;
            $g \gets 0$ (Initialize the generation counter)\;
            $archive \gets$ Initialize to empty set\;
		\While{ $g \leq N_{gen}$ }{
			\For{each individual architecture ($\alpha$) in $\mathcal{P}$} {
                    $F_{rs} \gets \phi(\alpha^*, \alpha)$ (using Equation~\ref{eq:rmi})\;\label{line:RS}
                    $F_{hw} \gets \Psi(\alpha)$ \;\label{line:HW}
                    $F_{div} \gets \chi(\alpha, \mathcal{P})$ (using Equation~\ref{eq:hardware_diversity})\;\label{line:DIV}
			}
                UpdateArchive($\mathcal{P}$, $archive$)\;\label{line:UpdateArchive}
			$g \gets g + 1$\;
                $\mathcal{P} \gets $ NSGA-II($F_{rs}$, $F_{hw}$, $F_{div}$)\;\label{line:NSGA}
		}		
    \end{algorithm}

\begin{figure*}[t]
    \centering
    \includegraphics[width=\linewidth]{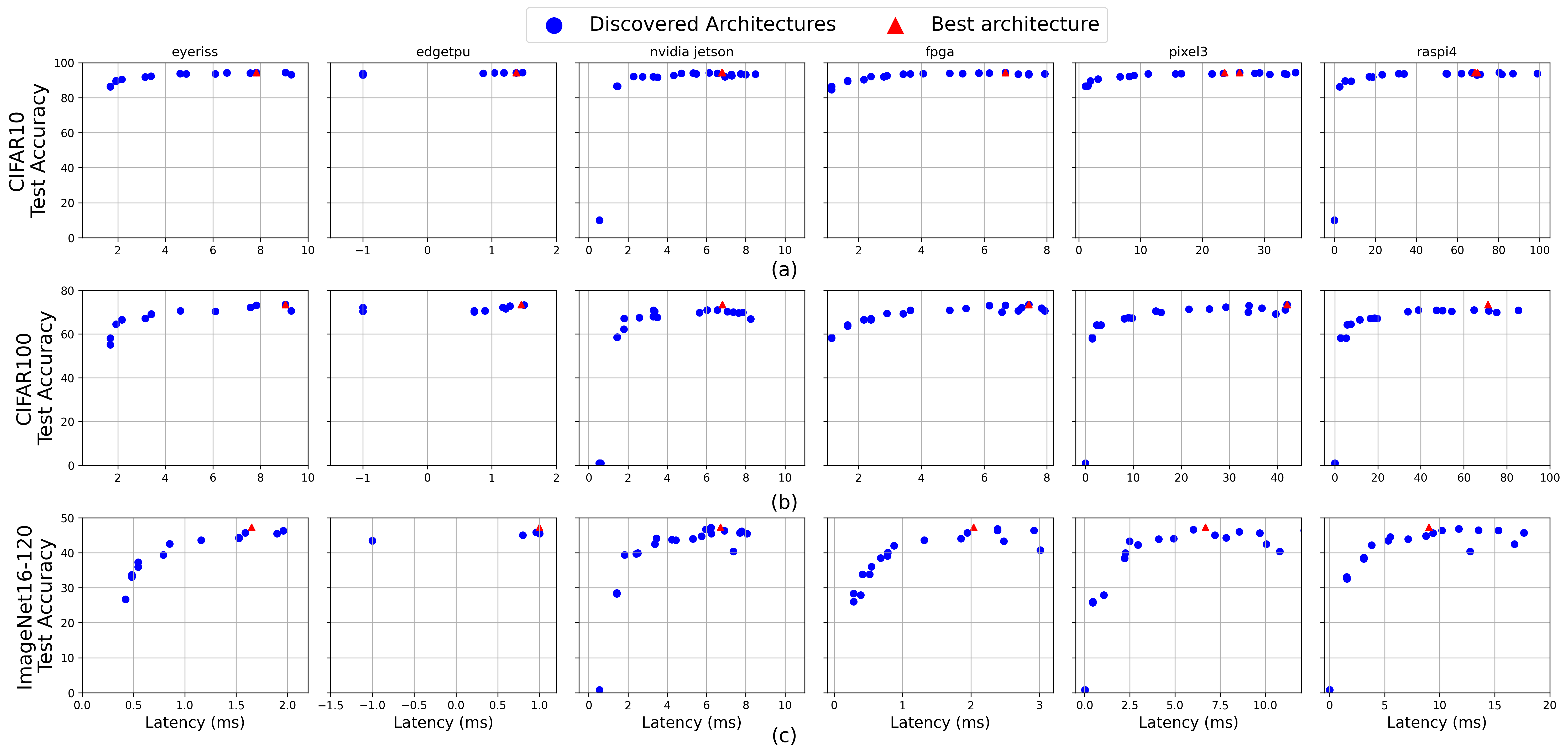}
    \caption{Results of \methodname$\ $ for 6 different edge devices performed with only 3 objectives: maximize representation similarity, minimizing device latency and maximizing the hardware cost diversity.
    (a), (b), (c) show the results for image classification task on CIFAR10, CIFAR100 and ImageNet16-120 respectively.}
    \label{fig:3objs_res}
\end{figure*}
\subsection{\methodname}
\label{subsect:method}    
    The pseudo-code of the proposed \methodname$\ $is presented in Algorithm~\ref{algo}.
    It begins by initializing a population $\mathcal{P}$ consisting of $N_{pop}$ architectures randomly  sampled from $\mathcal{A}$.
    \methodname$\ $iterates for $N_{gen}$ generations.
    During each generation, the performance of every architecture $\alpha$ ($F_{rs}$) in the current population, in terms of similarity to reference model, is evaluated in \textit{line}~\ref{line:RS} using Equation~\ref{eq:rmi}. Next, hardware cost $F_{hw}$ and hardware cost diversity $F_{div}$ of the architectures are calculated in \textit{lines}~\ref{line:HW}, \ref{line:DIV} respectively.
    Then, the archive is updated  in \textit{line}~\ref{line:UpdateArchive} to include the new architectures from the current population.
    Finally, NSGA-II is used to generate the next generation population in \textit{line}~\ref{line:NSGA}.
    \methodname$\ $ returns a pareto optimal front of architectures $P_{optimal}$ (\ie set of possible neural architecture solutions) after $N_{gen}$ generations.
\section{Experiments}
\label{section:experiments}
We adopt the architecture representation introduced in \cite{sinha2021evolving} and conduct the architecture search using a single NVIDIA RTX A4000 GPU, with a population size ($N_{pop}$) set to 20. 
Following \cite{zheng2022neural}, we employ ResNet-20 as  the reference model.
The representation similarity score is calculated in accordance to the procedure outlined in~\cite{sinha2024hardware} and the architecture search is performed for 100 generations ($N_{gen}$).


Further details on the experiments, such as the search space and datasets are presented in Section~\ref{subsection:search_space} and Section~\ref{subsection:datasets}, respectively. 
Section~\ref{subsection:results} reports the architecture search performance for six different edge devices, considering various hardware cost settings for each. Finally, an ablation study is performed on the hardware cost diversity objective in Section~\ref{subsection:ablation}.

\subsection{Search Space}
\label{subsection:search_space}
    The effectiveness of the proposed method is demonstrated on the \textbf{NAS-Bench-201}~\cite{Dong2020NAS-Bench-201} benchmark search space. 
    It provides a unified benchmark for fair comparison of NAS algorithms by providing the results on CIFAR-10, CIFAR-100 and ImageNet16-120 for image classification task.
    Given that any NAS algorithm aims to search for the type of the operation present between two nodes in a neural architecture, the search space of NAS-Bench-201 includes convolution 3x3, convolution 1x1, max pooling 3x3, skip connection, and none.
    Note that \textit{none} indicates the absence of any operation between the two nodes.
    Nevertheless, NAS-Bench-201 lacks information about the hardware cost associated with its architectures.
    Consequently, we utilize the \textbf{HW-NAS-Bench} \cite{li2021hwnasbench} benchmark.
    It is an extension of NAS-Bench-201, containing various hardware costs for all architectures in its search space across six edge devices including, \textit{NVIDIA Edge GPU Jetson TX2, Raspberry Pi 4, Edge TPU, Pixel 3, ASIC-Eyeriss, and FPGA}.

\begin{figure}[t]
    \begin{minipage}{0.45\textwidth}
        \centering
        \includegraphics[width=\linewidth]{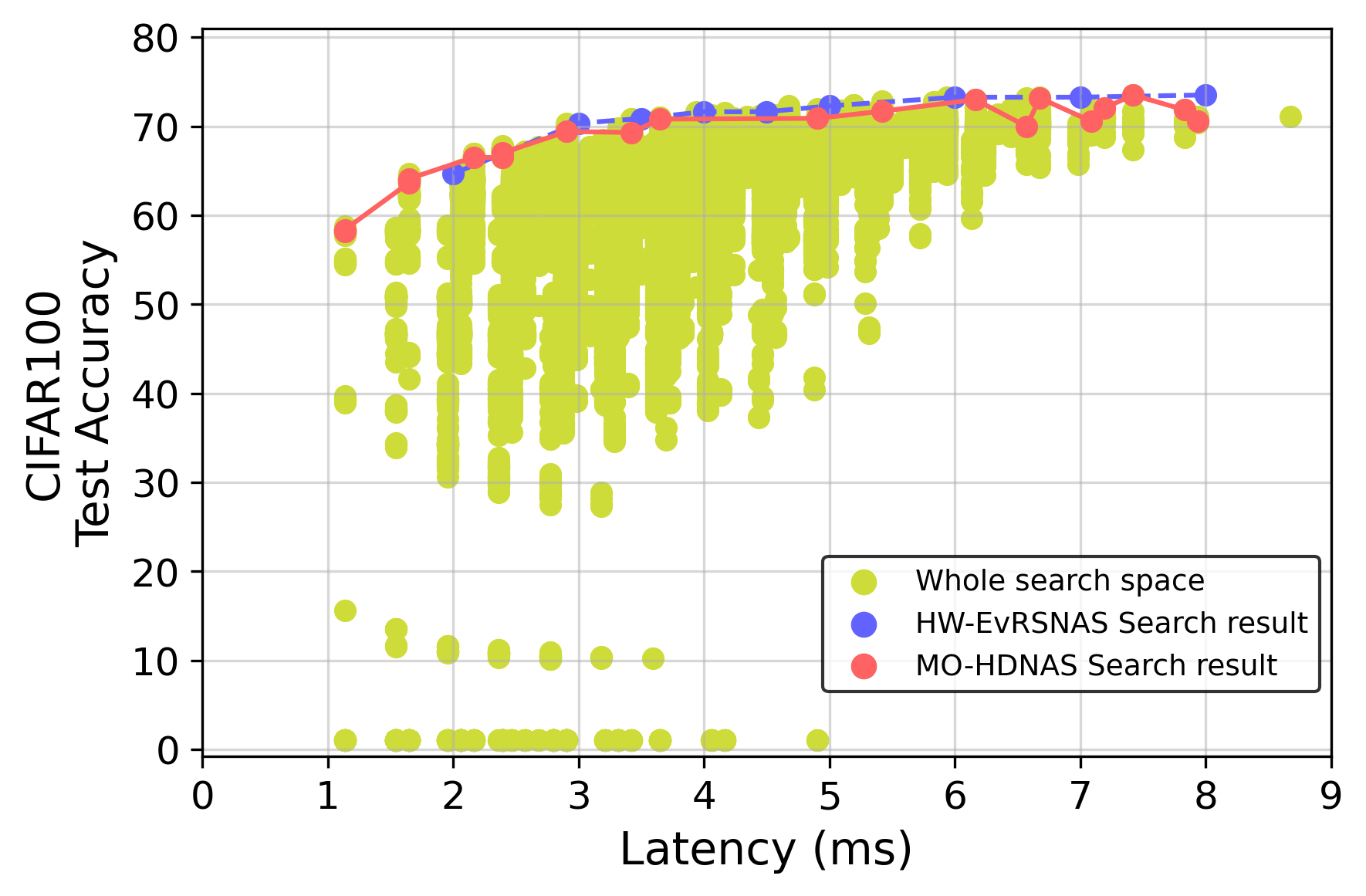}
        \captionof{figure}{Comparison of architecture search results for FPGA on CIFAR-100 dataset between HW-EvRSNAS~\cite{sinha2024hardware} and \methodname.}
        \label{fig:single_vs_multi}
    \end{minipage}%
    \hfill
    \begin{minipage}{0.45\textwidth}
    \centering
        \begin{tabular}{|c|c|}
            \hline
                                                & Search cost \\
                            Methods             & (GPU hours) \\
            \hline
            HW-EvRSNAS~\cite{sinha2024hardware} & $20.87$ \\
            \textbf{\methodname$\ $(Ours) }     & \bf{$0.65$} \\
            \hline
        \end{tabular}
    \captionof{table}{Search cost comparison of architecture search results for FPGA on CIFAR-100 dataset between HW-EvRSNAS~\cite{sinha2024hardware} and our method.}
    \label{tab:single_vs_multi}
    \end{minipage}
\end{figure}

\begin{figure*}[t]
    \centering
    \includegraphics[width=\linewidth]{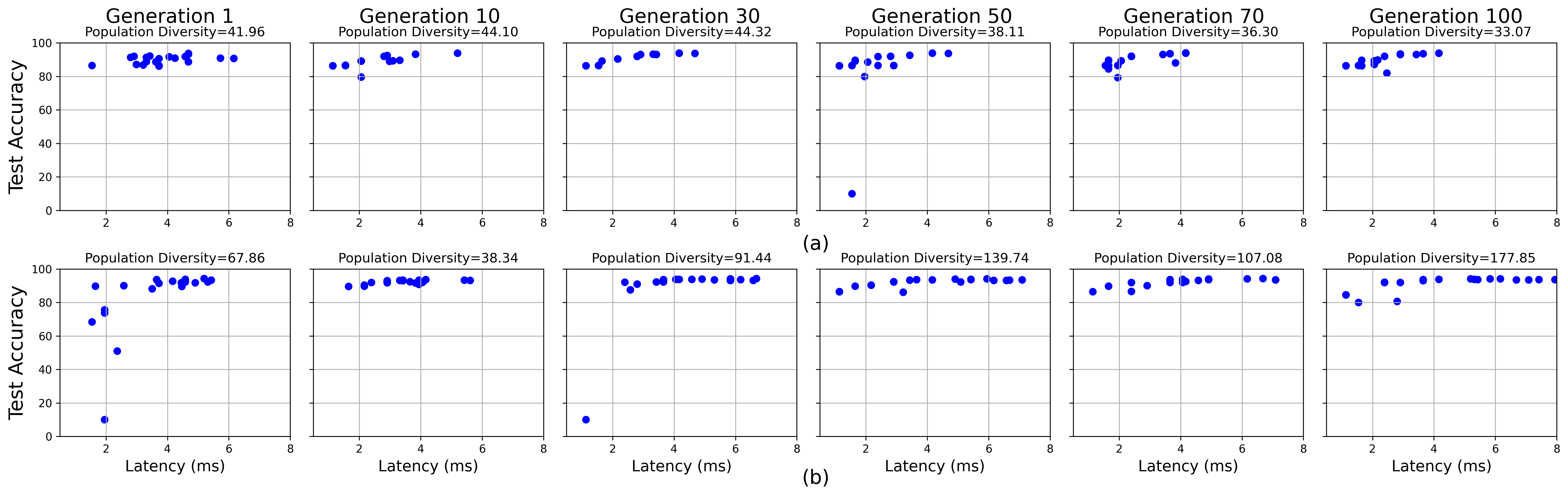}
    \caption{The average hardware costs diversity across different generation population for (a) two objectives (b) three objectives.
    Note that  the search was conducted for the FPGA device using the CIFAR-10 dataset.
    The x-axis represents the architecture latency on FPGA, while the y-axis depicts the test accuracy of the architecture on CIFAR-10.}
    \label{fig:com_diversity_fpga_c10}
\end{figure*}
\subsection{Datasets}
\label{subsection:datasets}
    We test the effectiveness of the proposed method on three different datasets: \textit{CIFAR-10}, \textit{CIFAR-100}, and \textit{ImageNet-16-120}. \textit{CIFAR-10} consists of 50,000 train and 10,000 test images, categorized into 10 classes. As for the \textit{CIFAR-100}, the number of images in the train and test sets are the same as \textit{CIFAR-10}, but instead coming from 100 classes. 
    On the other hand, the \textit{ImageNet-16-120} \cite{chrabaszcz2017downsampled} is a modified version of ImageNet containing  120 out of the 1000 total labels, and with each image being downsampled to $16\times16$ pixels.


\subsection{Results}
\label{subsection:results}
    Architecture search results obtained by the proposed \methodname$\ $method are shown in Figure~\ref{fig:3objs_res}.
    It shows the pareto fronts for the multi-objective architecture search performed on six different edge devices and their hardware cost measured in terms of latency.
    It is observed that the architectures present in the pareto front of \methodname$\ $are closer to the best architecture.
    Notably, the pareto front contains architectures with diverse latencies.

    To further evaluate the effectiveness of our proposed method, we compare our architecture search results with those of HW-EvRSNAS~\cite{sinha2024hardware}, which treats the HW-NAS problem as a single objective optimization (Figure~\ref{fig:proposed_method}).
    Results are illustrated in Figure~\ref{fig:single_vs_multi}, presenting the pareto front discovered by \methodname$\ $ and the architecture search results of HW-EvRSNAS under nine different hardware cost constraints.
    Note that these results are obtained for the image classification task on the CIFAR100 dataset using FPGA.
    From the figure, it is evident that our method is able to identify a more diverse set of high-performing architectures, ranging from those with low latency to those with high latency.

    Furthermore, We compare the search cost of our method with that of HW-EvRSNAS~\cite{sinha2024hardware} in Table~\ref{tab:single_vs_multi}.
    Search costs are reported in terms of GPU hours, indicating the number of hours each method spent to perform the architecture search on a single GPU.
    These results demonstrate that our method finds the pareto set of architectures at a search cost that is $32\times$ lower than that of HW-EvRSNAS.
    This is attributed to the fact that HW-EvRSNAS requires nine separate runs to find the optimal architecture for nine different hardware cost constraints. In contrast, our method finds the pareto set of 20 architectures in just a single run.

\subsection{Ablation Study}
\label{subsection:ablation}
    To illustrate the influence of the third objective in the Eq~\ref{eq:problem} (\ie maximizing hardware cost diversity $\chi(\alpha, \mathcal{P})$), we visualize the \textit{population diversity}, $\bar{\chi}(\mathcal{P})$, 
    of 6 different generations (\textit{1, 10, 30, 50, 70, 100}) in Figure~\ref{fig:com_diversity_fpga_c10}.
    $\bar{\chi}(\mathcal{P})$ measures the average hardware cost diversity for a generation, as showcased in Equation~\ref{eq:avg_hardware_diversity}.
    It is computed by taking the average of the $\chi(\alpha, \mathcal{P})$ term across all architectures within that generation's population.
    Figure~\ref{fig:com_diversity_fpga_c10}(a) illustrates the progression of the population diversity term across generations when the objective of maximizing hardware cost diversity is not applied in Equation~\ref{eq:problem} (\ie only the first two objectives employed).
    In this case, we observe a decline in  population diversity as generations progress.
    This results in architectures within the population being inclined towards regions with high accuracy and lower hardware costs.
    Consequently, it hinders the discovery of the best architecture within the search area characterized by high hardware cost.
    
    On the other hand, Figure~\ref{fig:com_diversity_fpga_c10}(b) shows the progression of the population diversity term across generations when all three objectives in Equation~\ref{eq:problem} are utilized in the search process.
    In this scenario, we observe an increase in population diversity as generation advances.
    Hence, architectures within the population exhibit diverse hardware costs, spanning from low to high latencies.
    This enhances the explorability of the search process, facilitating the discovery of high-performing architectures within regions characterized by high hardware cost.
\section{Conclusion}
\label{sec:conclusion}
    In this work, we presented a multi-objective hardware aware neural architecture search method, which performs the architecture search with reduced computational cost. This is achieved by searching for architectures with similar internal representation to a reference model, and simultaneously, with minimum hardware cost.
    Additionally, we introduced a third search objective, hardware cost diversity, to facilitate a better exploration of the architecture search space.
    The effectiveness of the proposed method is demonstrated on six edge devices for image classification task on three different datasets.

\section{Acknowledgement}
    This work is supported by the Luxembourg National Research Fund (FNR), under the project reference C21/IS/15965298/ELITE.
    
{
    \small
    \bibliographystyle{ieeenat_fullname}
    \bibliography{references}

\begin{thebibliography}{37}
\providecommand{\natexlab}[1]{#1}
\providecommand{\url}[1]{\texttt{#1}}
\expandafter\ifx\csname urlstyle\endcsname\relax
  \providecommand{\doi}[1]{doi: #1}\else
  \providecommand{\doi}{doi: \begingroup \urlstyle{rm}\Url}\fi

\bibitem[Benmeziane et~al.(2021{\natexlab{a}})Benmeziane, El~Maghraoui, Ouarnoughi, Niar, Wistuba, and Wang]{ijcai2021p592}
Hadjer Benmeziane, Kaoutar El~Maghraoui, Hamza Ouarnoughi, Smail Niar, Martin Wistuba, and Naigang Wang.
\newblock Hardware-aware neural architecture search: Survey and taxonomy.
\newblock In \emph{Proceedings of the Thirtieth International Joint Conference on Artificial Intelligence, {IJCAI-21}}, pages 4322--4329. International Joint Conferences on Artificial Intelligence Organization, 2021{\natexlab{a}}.
\newblock Survey Track.

\bibitem[Benmeziane et~al.(2021{\natexlab{b}})Benmeziane, Maghraoui, Ouarnoughi, Niar, Wistuba, and Wang]{benmeziane2021comprehensive}
Hadjer Benmeziane, Kaoutar~El Maghraoui, Hamza Ouarnoughi, Smail Niar, Martin Wistuba, and Naigang Wang.
\newblock A comprehensive survey on hardware-aware neural architecture search.
\newblock \emph{arXiv preprint arXiv:2101.09336}, 2021{\natexlab{b}}.

\bibitem[Cai et~al.(2019{\natexlab{a}})Cai, Gan, Wang, Zhang, and Han]{cai2019once}
Han Cai, Chuang Gan, Tianzhe Wang, Zhekai Zhang, and Song Han.
\newblock Once-for-all: Train one network and specialize it for efficient deployment.
\newblock \emph{arXiv preprint arXiv:1908.09791}, 2019{\natexlab{a}}.

\bibitem[Cai et~al.(2019{\natexlab{b}})Cai, Zhu, and Han]{cai2018proxylessnas}
Han Cai, Ligeng Zhu, and Song Han.
\newblock Proxyless{NAS}: direct neural architecture search on target task and hardware.
\newblock In \emph{International Conference on Learning Representations}, 2019{\natexlab{b}}.

\bibitem[Chrabaszcz et~al.(2017)Chrabaszcz, Loshchilov, and Hutter]{chrabaszcz2017downsampled}
Patryk Chrabaszcz, Ilya Loshchilov, and Frank Hutter.
\newblock A downsampled variant of imagenet as an alternative to the cifar datasets.
\newblock \emph{arXiv preprint arXiv:1707.08819}, 2017.

\bibitem[Chu et~al.(2020)Chu, Zhang, and Xu]{chu2020multi}
Xiangxiang Chu, Bo Zhang, and Ruijun Xu.
\newblock Multi-objective reinforced evolution in mobile neural architecture search.
\newblock In \emph{European Conference on Computer Vision}, pages 99--113. Springer, 2020.

\bibitem[Collobert et~al.(2011)Collobert, Weston, Bottou, Karlen, Kavukcuoglu, and Kuksa]{collobert2011natural}
Ronan Collobert, Jason Weston, L{\'e}on Bottou, Michael Karlen, Koray Kavukcuoglu, and Pavel Kuksa.
\newblock Natural language processing (almost) from scratch.
\newblock \emph{Journal of machine learning research}, 12\penalty0 (ARTICLE):\penalty0 2493--2537, 2011.

\bibitem[Deb et~al.(2002)Deb, Pratap, Agarwal, and Meyarivan]{deb2002fast}
Kalyanmoy Deb, Amrit Pratap, Sameer Agarwal, and TAMT Meyarivan.
\newblock A fast and elitist multiobjective genetic algorithm: Nsga-ii.
\newblock \emph{IEEE transactions on evolutionary computation}, 6\penalty0 (2):\penalty0 182--197, 2002.

\bibitem[Devlin et~al.(2018)Devlin, Chang, Lee, and Toutanova]{devlin2018bert}
Jacob Devlin, Ming-Wei Chang, Kenton Lee, and Kristina Toutanova.
\newblock Bert: Pre-training of deep bidirectional transformers for language understanding.
\newblock \emph{arXiv preprint arXiv:1810.04805}, 2018.

\bibitem[Dong and Yang(2020)]{Dong2020NAS-Bench-201}
Xuanyi Dong and Yi Yang.
\newblock Nas-bench-201: Extending the scope of reproducible neural architecture search.
\newblock In \emph{International Conference on Learning Representations}, 2020.

\bibitem[Elsken et~al.(2018)Elsken, Metzen, and Hutter]{elsken2018neural}
Thomas Elsken, Jan~Hendrik Metzen, and Frank Hutter.
\newblock Neural architecture search: A survey.
\newblock \emph{arXiv preprint arXiv:1808.05377}, 2018.

\bibitem[Elsken et~al.(2019)Elsken, Metzen, Hutter, et~al.]{elsken2019neural}
Thomas Elsken, Jan~Hendrik Metzen, Frank Hutter, et~al.
\newblock Neural architecture search: A survey.
\newblock \emph{J. Mach. Learn. Res.}, 20\penalty0 (55):\penalty0 1--21, 2019.

\bibitem[Garcia et~al.(2021)Garcia, Musallam, Gaudilliere, Ghorbel, Al~Ismaeil, Perez, and Aouada]{garcia2021lspnet}
Albert Garcia, Mohamed~Adel Musallam, Vincent Gaudilliere, Enjie Ghorbel, Kassem Al~Ismaeil, Marcos Perez, and Djamila Aouada.
\newblock Lspnet: A 2d localization-oriented spacecraft pose estimation neural network.
\newblock In \emph{Proceedings of the IEEE/CVF Conference on Computer Vision and Pattern Recognition}, pages 2048--2056, 2021.

\bibitem[Goldberg(2013)]{goldberg2013genetic}
David~E Goldberg.
\newblock \emph{Genetic algorithms}.
\newblock pearson education India, 2013.

\bibitem[Kornblith et~al.(2019)Kornblith, Norouzi, Lee, and Hinton]{kornblith2019similarity}
Simon Kornblith, Mohammad Norouzi, Honglak Lee, and Geoffrey Hinton.
\newblock Similarity of neural network representations revisited.
\newblock In \emph{International conference on machine learning}, pages 3519--3529. PMLR, 2019.

\bibitem[Li et~al.(2021)Li, Yu, Fu, Zhang, Zhao, You, Yu, Wang, and Lin]{li2021hwnasbench}
Chaojian Li, Zhongzhi Yu, Yonggan Fu, Yongan Zhang, Yang Zhao, Haoran You, Qixuan Yu, Yue Wang, and Yingyan~(Celine) Lin.
\newblock {\{}HW{\}}-{\{}nas{\}}-bench: Hardware-aware neural architecture search benchmark.
\newblock In \emph{International Conference on Learning Representations}, 2021.

\bibitem[Liu et~al.(2019)Liu, Simonyan, and Yang]{liu2018darts2}
Hanxiao Liu, Karen Simonyan, and Yiming Yang.
\newblock {DARTS}: Differentiable architecture search.
\newblock In \emph{International Conference on Learning Representations}, 2019.

\bibitem[Lu et~al.(2019)Lu, Whalen, Boddeti, Dhebar, Deb, Goodman, and Banzhaf]{lu2019nsga}
Zhichao Lu, Ian Whalen, Vishnu Boddeti, Yashesh Dhebar, Kalyanmoy Deb, Erik Goodman, and Wolfgang Banzhaf.
\newblock Nsga-net: neural architecture search using multi-objective genetic algorithm.
\newblock In \emph{Proceedings of the Genetic and Evolutionary Computation Conference}, pages 419--427, 2019.

\bibitem[Musallam et~al.(2021)Musallam, Gaudilliere, Ghorbel, Ismaeil, Perez, Poucet, and Aouada]{9620184}
Mohamed~Adel Musallam, Vincent Gaudilliere, Enjie Ghorbel, Kassem~Al Ismaeil, Marcos~Damian Perez, Michel Poucet, and Djamila Aouada.
\newblock Spacecraft recognition leveraging knowledge of space environment: Simulator, dataset, competition design and analysis.
\newblock In \emph{2021 IEEE International Conference on Image Processing Challenges (ICIPC)}, pages 11--15, 2021.

\bibitem[Perez et~al.(2021)Perez, Mohamed~Ali, Garcia~Sanchez, Ghorbel, Al~Ismaeil, Le~Henaff, and Aouada]{perez2021detection}
Marcos Perez, Mohamed~Adel Mohamed~Ali, Albert Garcia~Sanchez, Enjie Ghorbel, Kassem Al~Ismaeil, Paul Le~Henaff, and Djamila Aouada.
\newblock Detection \& identification of on-orbit objects using machine learning.
\newblock In \emph{European Conference on Space Debris}, 2021.

\bibitem[Pham et~al.(2018)Pham, Guan, Zoph, Le, and Dean]{pmlr-v80-pham18a}
Hieu Pham, Melody Guan, Barret Zoph, Quoc Le, and Jeff Dean.
\newblock Efficient neural architecture search via parameters sharing.
\newblock In \emph{Proceedings of the 35th International Conference on Machine Learning}, pages 4095--4104, Stockholmsmässan, Stockholm Sweden, 2018. PMLR.

\bibitem[Real et~al.(2019)Real, Aggarwal, Huang, and Le]{real2019regularized}
Esteban Real, Alok Aggarwal, Yanping Huang, and Quoc~V Le.
\newblock Regularized evolution for image classifier architecture search.
\newblock In \emph{Proceedings of the AAAI Conference on Artificial Intelligence}, pages 4780--4789, 2019.

\bibitem[Rostami et~al.(2022)Rostami, Zamani, Fakharzadeh, Amini, and Marvasti]{rostami2022deep}
Peyman Rostami, Hojatollah Zamani, Mohammad Fakharzadeh, Arash Amini, and Farokh Marvasti.
\newblock A deep learning approach for reconstruction in millimeter-wave imaging systems.
\newblock \emph{IEEE Transactions on Antennas and Propagation}, 71\penalty0 (1):\penalty0 1180--1184, 2022.

\bibitem[Sermanet et~al.(2013)Sermanet, Eigen, Zhang, Mathieu, Fergus, and LeCun]{sermanet2013overfeat}
Pierre Sermanet, David Eigen, Xiang Zhang, Micha{\"e}l Mathieu, Rob Fergus, and Yann LeCun.
\newblock Overfeat: Integrated recognition, localization and detection using convolutional networks.
\newblock \emph{arXiv preprint arXiv:1312.6229}, 2013.

\bibitem[Simonyan and Zisserman(2014)]{simonyan2014very}
Karen Simonyan and Andrew Zisserman.
\newblock Very deep convolutional networks for large-scale image recognition.
\newblock \emph{arXiv preprint arXiv:1409.1556}, 2014.

\bibitem[Sinha and Chen(2021)]{sinha2021evolving}
Nilotpal Sinha and Kuan-Wen Chen.
\newblock Evolving neural architecture using one shot model.
\newblock In \emph{Proceedings of the Genetic and Evolutionary Computation Conference}, pages 910--918, 2021.

\bibitem[Sinha and Chen(2022{\natexlab{a}})]{sinha2022neural}
Nilotpal Sinha and Kuan-Wen Chen.
\newblock Neural architecture search using progressive evolution.
\newblock In \emph{Proceedings of the Genetic and Evolutionary Computation Conference}, pages 1093--1101, 2022{\natexlab{a}}.

\bibitem[Sinha and Chen(2022{\natexlab{b}})]{sinha2022novelty}
Nilotpal Sinha and Kuan-Wen Chen.
\newblock Novelty driven evolutionary neural architecture search.
\newblock In \emph{Proceedings of the Genetic and Evolutionary Computation Conference Companion}, pages 671--674, 2022{\natexlab{b}}.

\bibitem[Sinha and Chen(2023)]{10.1162/evco_a_00331}
Nilotpal Sinha and Kuan-Wen Chen.
\newblock {Neural Architecture Search Using Covariance Matrix Adaptation Evolution Strategy}.
\newblock \emph{Evolutionary Computation}, pages 1--28, 2023.

\bibitem[Sinha et~al.(2024)Sinha, El~Rahman~Shabayek, Kacem, Rostami, Shneider, and Aouada]{sinha2024hardware}
Nilotpal Sinha, Abd El~Rahman~Shabayek, Anis Kacem, Peyman Rostami, Carl Shneider, and Djamila Aouada.
\newblock Hardware aware evolutionary neural architecture search using representation similarity metric.
\newblock In \emph{Proceedings of the IEEE/CVF Winter Conference on Applications of Computer Vision}, pages 2628--2637, 2024.

\bibitem[Wu et~al.(2016)Wu, Schuster, Chen, Le, Norouzi, Macherey, Krikun, Cao, Gao, Macherey, et~al.]{wu2016google}
Yonghui Wu, Mike Schuster, Zhifeng Chen, Quoc~V Le, Mohammad Norouzi, Wolfgang Macherey, Maxim Krikun, Yuan Cao, Qin Gao, Klaus Macherey, et~al.
\newblock Google's neural machine translation system: Bridging the gap between human and machine translation.
\newblock \emph{arXiv preprint arXiv:1609.08144}, 2016.

\bibitem[Ying et~al.(2020)Ying, Zheng, Wu, Li, and Xu]{ying2020neural}
Weiqin Ying, Kaijie Zheng, Yu Wu, Junhui Li, and Xin Xu.
\newblock Neural architecture search using multi-objective evolutionary algorithm based on decomposition.
\newblock In \emph{Artificial Intelligence Algorithms and Applications: 11th International Symposium, ISICA 2019, Guangzhou, China, November 16--17, 2019, Revised Selected Papers 11}, pages 143--154. Springer, 2020.

\bibitem[Zamani et~al.(2021)Zamani, Rostami, Amini, and Marvasti]{zamani2021elliptical}
Hojatollah Zamani, Peyman Rostami, Arash Amini, and Farokh Marvasti.
\newblock Elliptical shape recovery from blurred pixels using deep learning.
\newblock In \emph{ICASSP 2021-2021 IEEE International Conference on Acoustics, Speech and Signal Processing (ICASSP)}, pages 2775--2779. IEEE, 2021.

\bibitem[Zela et~al.(2020)Zela, Elsken, Saikia, Marrakchi, Brox, and Hutter]{Zela2020Understanding}
Arber Zela, Thomas Elsken, Tonmoy Saikia, Yassine Marrakchi, Thomas Brox, and Frank Hutter.
\newblock Understanding and robustifying differentiable architecture search.
\newblock In \emph{International Conference on Learning Representations}, 2020.

\bibitem[Zheng et~al.(2022)Zheng, Fei, Zhang, Wu, Chao, Liu, Zeng, Tian, and Ji]{zheng2022neural}
Xiawu Zheng, Xiang Fei, Lei Zhang, Chenglin Wu, Fei Chao, Jianzhuang Liu, Wei Zeng, Yonghong Tian, and Rongrong Ji.
\newblock Neural architecture search with representation mutual information.
\newblock In \emph{Proceedings of the IEEE/CVF Conference on Computer Vision and Pattern Recognition}, pages 11912--11921, 2022.

\bibitem[Zoph and Le(2017)]{zoph2017neural}
Barret Zoph and Quoc Le.
\newblock Neural architecture search with reinforcement learning.
\newblock In \emph{International Conference on Learning Representations}, 2017.

\bibitem[Zoph et~al.(2018)Zoph, Vasudevan, Shlens, and Le]{zoph2018learning}
Barret Zoph, Vijay Vasudevan, Jonathon Shlens, and Quoc~V Le.
\newblock Learning transferable architectures for scalable image recognition.
\newblock In \emph{Proceedings of the IEEE Conference on Computer Vision and Pattern Recognition}, pages 8697--8710, 2018.

\end{thebibliography}
}


\end{document}